\documentclass[runningheads]{llncs}
\usepackage{graphicx}

\usepackage{tikz}
\usepackage{comment} 
\usepackage{amsmath,amssymb} 
\usepackage{color}
\usepackage{caption}   
\usepackage{graphicx}  
\usepackage{caption}
\usepackage{subfigure} 
\usepackage{hyperref}

\begin{document}
\pagestyle{headings}
\mainmatter
\def\ECCVSubNumber{36}  

\title{The 1st Tiny Object Detection Challenge:\\ Methods and Results} 

\titlerunning{The 1st Tiny Object Detection Challenge: Methods and Results}
%
\author{Xuehui Yu\inst{1} \and Zhenjun Han\inst{1} \and Yuqi Gong\inst{1} \and Nan Jiang\inst{1} \and Jian Zhao\inst{2} \and Qixiang Ye\inst{1} \and
Jie Chen\inst{3} \and Yuan Feng\inst{4} \and Bin Zhang\inst{4} \and Xiaodi Wang\inst{4} \and Ying Xin\inst{4} \and Jingwei Liu\inst{5} \and Mingyuan Mao\inst{6} \and Sheng Xu\inst{6} \and Baochang Zhang\inst{6} \and Shumin Han\inst{4} \and Cheng Gao\inst{7} \and Wei Tang\inst{7} \and
Lizuo Jin\inst{7} \and Mingbo Hong\inst{8} \and Yuchao Yang\inst{8} \and Shuiwang Li\inst{8} \\
Huan Luo\inst{8} \and Qijun Zhao\inst{8} \and Humphrey Shi\inst{9}
}
\authorrunning{X. Yu et al.}
\institute{$^1$UCAS, $^2$NUS, $^3$Pengcheng Lab, $^4$Baidu Inc, $^5$ICT, CAS, $^6$Beihang University, $^7$Southeast University, $^8$Sichuan University, $^9$U of Oregon}
\maketitle

\begin{abstract}
The 1st Tiny Object Detection (TOD) Challenge aims to encourage research in developing novel and accurate methods for tiny object detection in images which have wide views, with a current focus on tiny person detection.  
The TinyPerson dataset was used for the TOD Challenge and is publicly released. It has 1610 images and 72651 box-level annotations. 
Around 36 participating teams from the globe competed in the 1st TOD Challenge.
In this paper, we provide a brief summary of the 1st TOD Challenge including brief introductions to the top three methods.The submission leaderboard will be reopened for researchers that are interested in the TOD challenge. The benchmark dataset and other information can be found at: \href{https://github.com/ucas-vg/TinyBenchmark}{\color{red} https://github.com/ucas-vg/TinyBenchmark}.

\end{abstract}

\section{Introduction}
Object detection is an important topic in the computer vision community. With the rise of deep convolutional neural networks, research in object detection has seen unprecedented progress~\cite{lin2014microsoft,ren2015faster,cheng2018revisiting,cheng2018decoupled,cai2018cascade,shen2017improving,zhang2019skynet}. Nevertheless, detecting tiny objects remains challenging and far from well-explored. One possible reason for this is because there is a lack of datasets and benchmarks for tiny object detection, and researchers thus pay much less attention to challenges in tiny object detection compared to general object detection.

Tiny object detection is very important for real-world vision application and differs from general object detection in several aspects. For example, since the objects are extremely small while the whole input image has relatively large field-of-view, there is much less information from the targeting objects and much more from background distractions. In addition, the large field-of-view characteristic of input images usually means that the tiny objects are imaged from a long distance, this makes detection of tiny objects with various poses and viewpoints even more difﬁcult. All these distinctions make tiny object detection a uniquely challenging task.


To encourage researchers to develop better methods to solve the tiny object detection problem, we held the first Tiny Object Detection Challenge. The TinyPerson dataset~\cite{yu2020scale} was adopted in this challenge. The dataset contains 1610 images with 72651 box-level annotations and is collected from real-world scenarios. The persons in the challenge dataset are very small, and their aspect ratios have a large variance so that they are representative of different type of objects. In the following section, we will summarize the detail information about the challenge, the methods and results.

\section{The TOD Challenge}

\subsection{Challenge Dataset}
\begin{figure}[htbp]
\subfigure[]{
\begin{minipage}[t]{0.5\linewidth}
\centering
\includegraphics[width=0.9\linewidth]{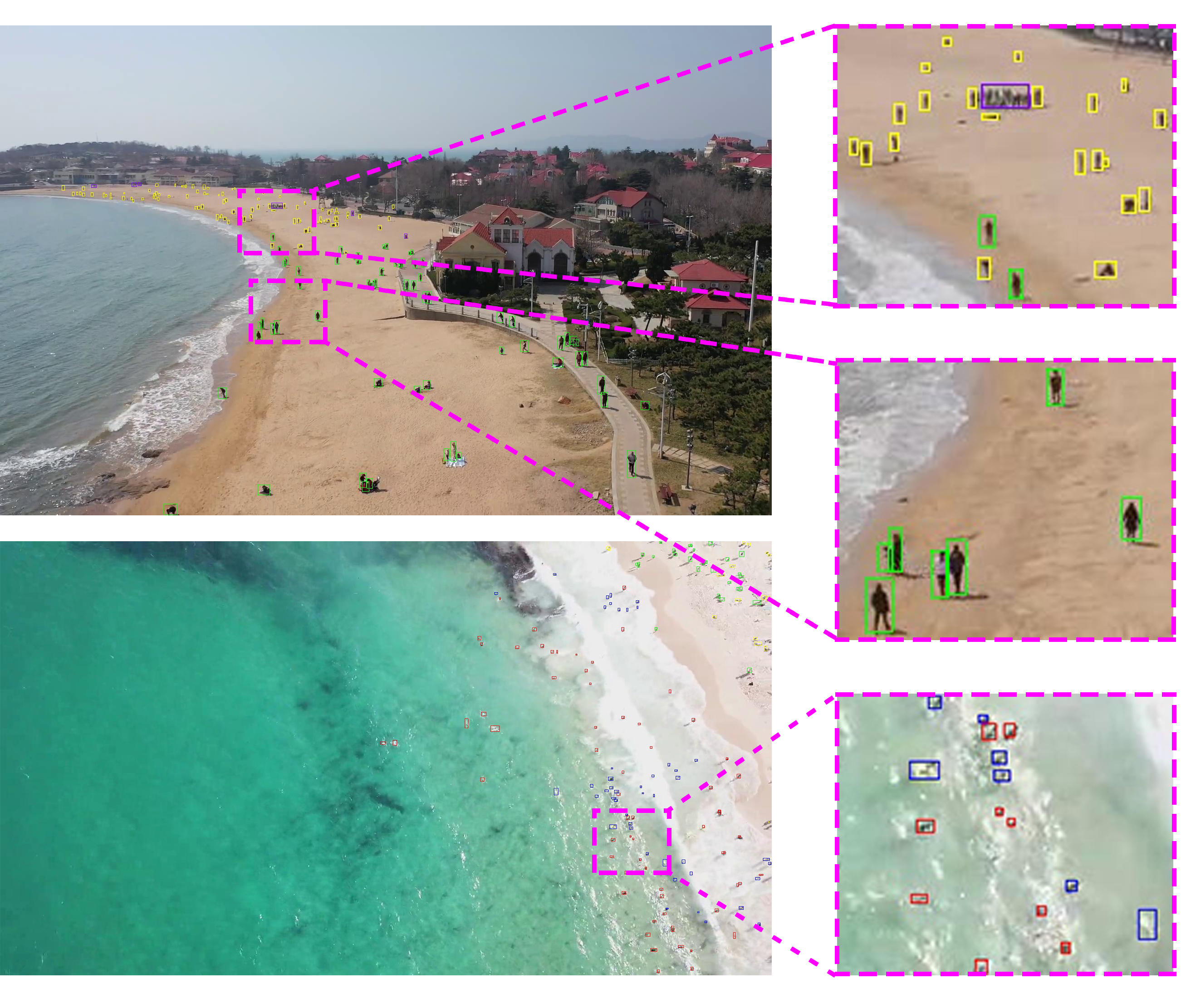}
\end{minipage}%
}
\subfigure[]{
\begin{minipage}[t]{0.4\linewidth}
\centering
   \includegraphics[width=0.9\linewidth]{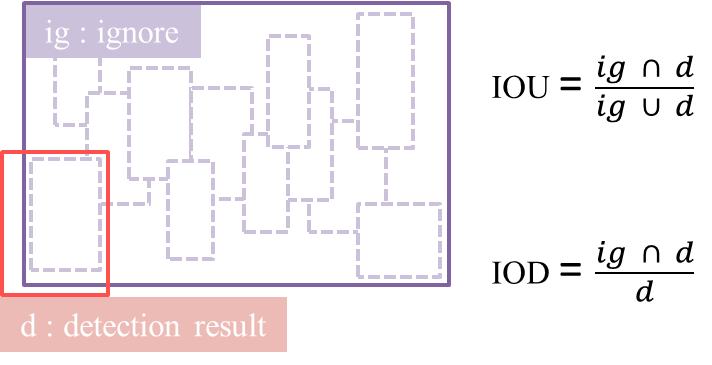}
\end{minipage}%
}%
\centering
\caption{(a): the annotation examples. “sea person”, “earth person”,“uncertain sea person”, “uncertain earth person”, "ignore region" are represented with red, green, blue, yellow, purple rectangle, respectively. The regions are zoomed in and shown on right. (b): IOU (insertion of union) and IOD (insertion of detection). IOD is for ignored regions for evaluation. The outline (inviolet) box represents a labeled ignored region and the dash boxes are unlabeled and ignored persons.  The red box is a detection’s result box that has high IOU with one ignored person.}
\label{Fig: annotations and IOD}
\end{figure}

The images used in the challenge are collected from some real-world videos to build the TinyPerson dataset~\cite{yu2020scale}. 
We sample images from video every 50 frames and delete images with a certain repetition for homogeneity. 
And finally, 72651 objects with bounding boxes are annotated. 
TinyPerson have four important properties: 
1) The persons in TinyPerson are quite tiny compared with other representative datasets, and the size of image are mainly 1920*1080, 
which is the main characteristics; 
2) The aspect ratio of persons in TinyPerson has a large variance. 
TinyPerson has the various poses and viewpoints of persons , it brings more complex diversity of the persons.
3) In TinyPerson, we mainly focus on person around seaside.
4) There are many images with dense objects (more than 200 persons per image) in TinyPerson.
In TinyPerson, we classify persons as “sea person” (persons in the sea) or “earth person” (persons on the land).
Some annotation rules in TinyPerson are deﬁned to determine which label a person belongs to(as an example show in Fig \ref{Fig: annotations and IOD}(a)): 
1) Persons on boat are treated as “sea person”; 
2) Persons lying in the water are treated as “sea person”; 
3) Persons with more than half body in water are treated as “sea person”; 
4) others are treated as “earth person”.
In addition, there are three conditions where persons are labeled as “ignore”: 
1) Crowds, which we can recognize as persons. But the crowds are hard to be separated one by one when labeled with standard rectangles;
2) Ambiguous regions, which are hard to clearly distinguish whether there is one or more persons, 
and 3) Reﬂections in water.Some objects are hard to be recognized as human beings, we directly labeled them as “uncertain”.

\subsection{Evaluation Metric}
We use both AP (average precision) and MR(miss rate) for performance evaluation.
The size range is divided into 3 intervals: tiny[2, 20], small[20, 32] and all[2, inf].
And for tiny[2, 20], it is partitioned into 3 sub-intervals: tiny1[2, 8], tiny2[8, 12], tiny3[12, 20].
And the IOU threshold is set to 0.25 and 0.5 for performance evaluation.
Same as pedestrian detection, 'ignore' regions do not participate in the evaluation, which means detection bounding boxes match them do not make a false positive. However in TinyPerson, most of ignore regions are much larger than that of a person. 
Therefore, we change IOU criteria to IOD(show in Fig 1 (b)) for ignore regions (IOD criteria only applies to ignore region, for other classes still use IOU criteria). In this challenge, we also treat uncertain same as ignore while evaluation.


\section{Result and Methods}
The 1st TOD challenge was held between April 20, 2020 and July 25, 2020. Around 36 teams submitted their final results in this challenge. Submission are evaluated on 786 images with 13787 person boxes and 1989 ignore regions, the images' resolution is mainly 1920*1080, some is even 3840*2160.
The results of the first TOD challenge show in Table \ref{Tab: challenge results}. In this section,we will briefly introduce the methodologies of the top 3 submissions.

\setlength{\tabcolsep}{4pt}
\begin{table}
\begin{center}
\caption{Challenge results ranked by $AP_{50}^{tiny}$}
\label{Tab: challenge results}
\begin{tabular}{llllll}
\hline\noalign{\smallskip}
Team          & $AP_{50}^{tiny}$ & $AP_{50}^{tiny1}$ & $AP_{50}^{tiny2}$ &$AP_{50}^{tiny3}$ & $AP_{25}^{tiny}$ \\
\noalign{\smallskip}
\hline
\noalign{\smallskip}
baidu\_ppdet  & 72.33 (1)  & 58.87 (8)   & 76.06 (1)   & 80.23 (1)   & 87.28 (1)  \\
pilafsama     & 71.53 (2)  & 59.21 (5)   & 75.22 (2)   & 79.68 (2)   & 85.27 (2)  \\
BingBing      & 71.36 (3)  & 59.91 (2)   & 74.70 (4)   & 78.63 (3)   & 84.74 (6)  \\
pplm          & 71.35 (4)  & 59.89 (4)   & 74.69 (5)   & 78.62 (4)   & 84.75 (5)  \\
tod666        & 71.34 (5)  & 59.97 (1)   & 74.70 (4)   & 78.57 (6)   & 84.61 (7)  \\
mix           & 71.32 (6)  & 59.90 (3)   & 74.68 (6)   & 78.60 (5)   & 84.61 (7)  \\
potting       & 70.91 (7)  & 58.87 (8)   & 74.73 (3)   & 78.18 (7)   & 84.93 (3)  \\
matter        & 70.10 (8)  & 58.88 (7)   & 73.06 (8)   & 77.49 (8)   & 84.13 (9)  \\
tiantian12414 & 69.71 (9)  & 57.88 (10)  & 73.67 (7)   & 77.26 (9)   & 84.25 (8)  \\
Mingbo\_Hong  & 69.34 (10) & 59.10 (6)   & 71.73 (11)  & 76.11 (13)  & 84.76 (4)  \\
dilidili      & 69.32 (11) & 58.28 (9)   & 72.71 (10)  & 76.55 (11)  & 83.38 (11) \\
LHX           & 69.20 (12) & 57.14 (11)  & 72.80 (9)   & 77.25 (10)  & 84.09 (10) \\
Washpan       & 68.73 (13) & 57.12 (12)  & 71.52 (12)  & 76.21 (12)  & 82.93 (12) \\
liw           & 67.87 (14) & 56.63 (13)  & 70.82 (13)  & 75.33 (14)  & 82.76 (13) \\
ZhangYuqi     & 65.31 (15) & 49.34 (20)  & 69.57 (15)  & 75.13 (15)  & 80.86 (17) \\
xieyy         & 65.27 (16) & 49.59 (18)  & 69.65 (14)  & 74.56 (16)  & 81.23 (16) \\
times         & 64.92 (17) & 49.49 (19)  & 69.37 (16)  & 73.98 (17)  & 81.58 (15) \\
Michealz      & 63.34 (18) & 53.55 (14)  & 65.58 (21)  & 69.66 (22)  & 81.77 (14) \\
yujia         & 62.94 (19) & 50.66 (16)  & 67.33 (19)  & 69.76 (21)  & 78.61 (21) \\
LLP           & 62.88 (20) & 50.97 (15)  & 66.64 (20)  & 69.78 (19)  & 79.58 (19) \\
ctt           & 62.83 (21) & 46.53 (21)  & 68.54 (17)  & 71.76 (18)  & 79.62 (18) \\
Lee\_Pisces   & 62.58 (22) & 50.55 (17)  & 67.59 (18)  & 69.77 (20)  & 78.93 (20) \\
xuesong       & 58.79 (23) & 44.81 (22)  & 61.46 (22)  & 68.81 (23)  & 76.73 (22) \\
alexto        & 57.52 (24) & 43.04 (23)  & 60.05 (23)  & 66.64 (24)  & 75.61 (23) \\
stevehsu      & 54.34 (25) & 35.74 (27)  & 59.04 (24)  & 65.94 (25)  & 74.91 (24) \\
fisty         & 52.88 (26) & 42.54 (24)  & 55.83 (25)  & 61.49 (29)  & 71.51 (26) \\
Evali         & 51.38 (27) & 37.06 (26)  & 55.09 (27)  & 62.51 (27)  & 72.87 (25) \\
mmeendez      & 51.07 (28) & 31.98 (28)  & 55.20 (26)  & 63.11 (26)  & 70.24 (28) \\
bobson        & 50.72 (29) & 38.82 (25)  & 53.73 (28)  & 58.84 (30)  & 70.52 (27) \\
daavoo        & 49.45 (30) & 30.36 (29)  & 52.78 (29)  & 61.52 (28)  & 68.08 (30) \\
divyanshahuja & 46.84 (31) & 29.77 (31)  & 51.96 (30)  & 57.74 (31)  & 68.46 (29) \\
xie233        & 44.67 (32) & 30.01 (30)  & 45.07 (31)  & 54.78 (32)  & 64.38 (31) \\
yingling      & 39.57 (33) & 24.30 (32)  & 43.98 (32)  & 50.80 (34)  & 61.91 (32) \\
zhaoxingjie   & 33.83 (34) & 5.09 (34)   & 34.17 (33)  & 52.91 (33)  & 57.49 (34) \\
suntinger     & 32.98 (35) & 16.88 (33)  & 32.11 (34)  & 47.43 (35)  & 60.30 (33) \\
Sugar\_bupt   & 13.61 (36) & 1.79 (35)   & 13.04 (35)  & 22.92 (36)  & 36.40 (35) \\
\hline
\end{tabular}
\end{center}
\end{table}
\setlength{\tabcolsep}{1.4pt}

\subsection{Team baidu$\_$ppdet}
\textbf{Yuan Feng, Bin Zhang, Xiaodi Wang, Ying Xin, Jingwei Liu,
Mingyuan Mao, Sheng Xu, Baochang Zhang, Shumin Han.}(Baidu \& Beihang University)\newline
\begin{figure}[t]
\begin{center}
   \includegraphics[width=0.7\linewidth]{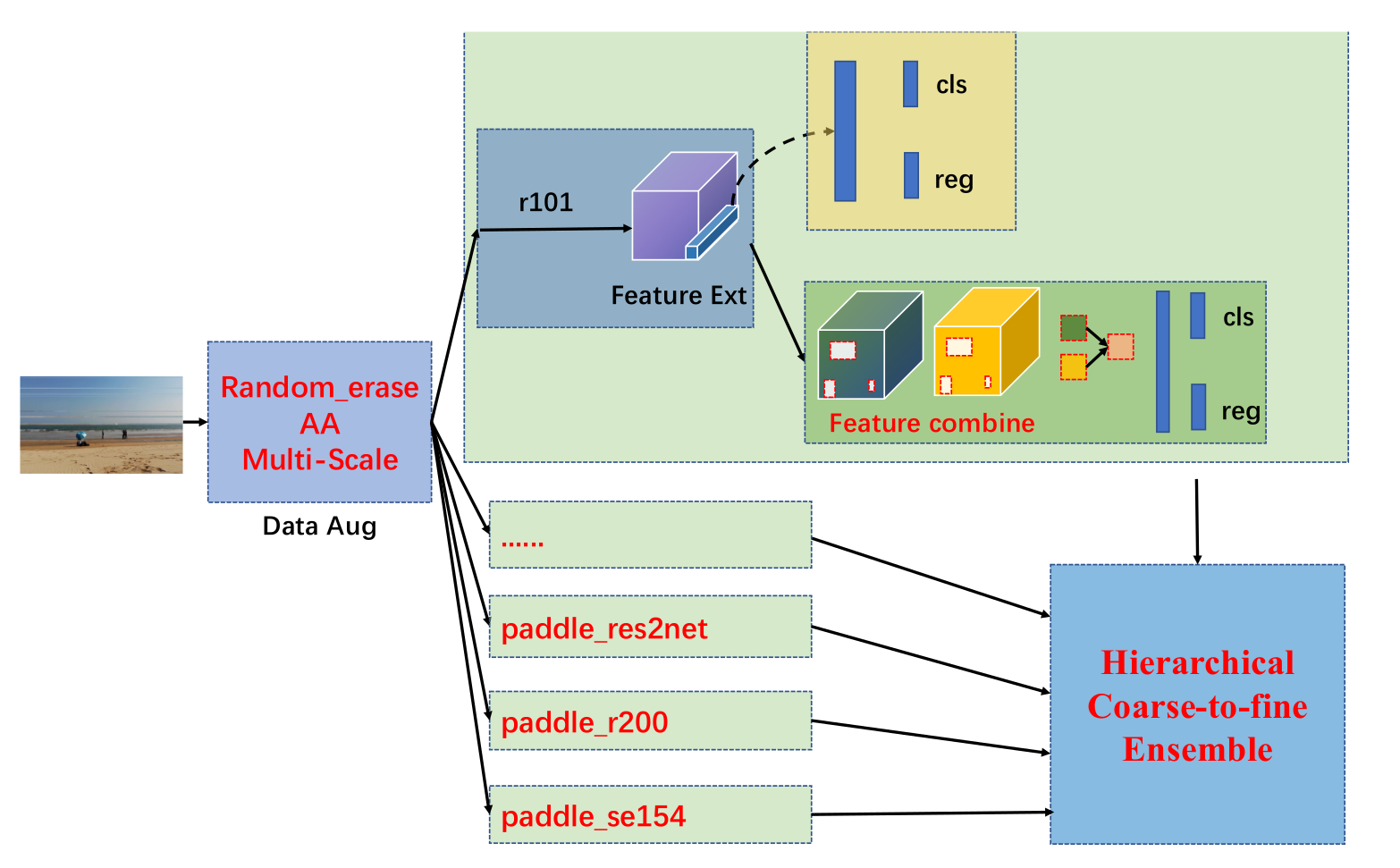}
\end{center}
   \caption{Framwork of Team baidu$\_$ppdet.}
\label{Fig:3.1 framework}
\end{figure}
Authors build detector based on the two-stage detection framework. They examine the performance of different components of detectors,leading to a large model pond for ensemble. The two-stage detectors include Faster R-CNN\cite{ren2015faster}, FPN\cite{lin2017feature}, Deformable R-CNN\cite{dai2017deformable}, and Cascade R-CNN\cite{cai2018cascade}. The training datasets are separated into two parts: 90$\%$ images for training and the remaining for validation. The framework is illustrated in Fig 2.\newline
\textbf{Data Augmentation.} Authors pretrained their models on MSCOCO\cite{lin2014microsoft} and Object365\cite{shao2019objects365} dataset for better performance.To address the scale variance issue, authors implement scale match\cite{yu2020scale} on MSCOCO by rescaling training images to match the size distribution of images in TinyPerson, which helps achieve 2$\%$ to 3$\%$ improvement in terms of AP 50$\%$.\newline
\textbf{Training Strategy.} Authors train Faster R-CNN with ResNet-101\cite{he2016deep} as baseline, and the AP$_{50}^{tiny}$ is 57.9$\%$. After applying multi-scale training tricks, the AP$_{50}^{tiny}$ reaches 60.1$\%$. Additionally,erase ignore regions while validate on valid set to keep same as evaluation also comes with earnings with nearly 2$\%$. Optimize the threshold of NMS, sample ratio and quantity. Finally, the AP$_{50}^{tiny}$ of FRCNN-Res101 reaches 65.38$\%$.\newline
\textbf{Model Refinement.} Feature representation is always the key to tiny object detection. A new feature fusion method is designed to improve the feature representation ability of networks. For Faster R-CNN and FPN, the P3 layer can better represent tiny object. In contrast, the lack of P3 layer brings the loss of the semantic information. Thus, authors use PAFPN\cite{tan2020efficientdet} to replace FPN in detector, which improves the mAP about 1.5$\%$.\newline 
\begin{figure}[t]
\begin{center}
   \includegraphics[width=1.0\linewidth]{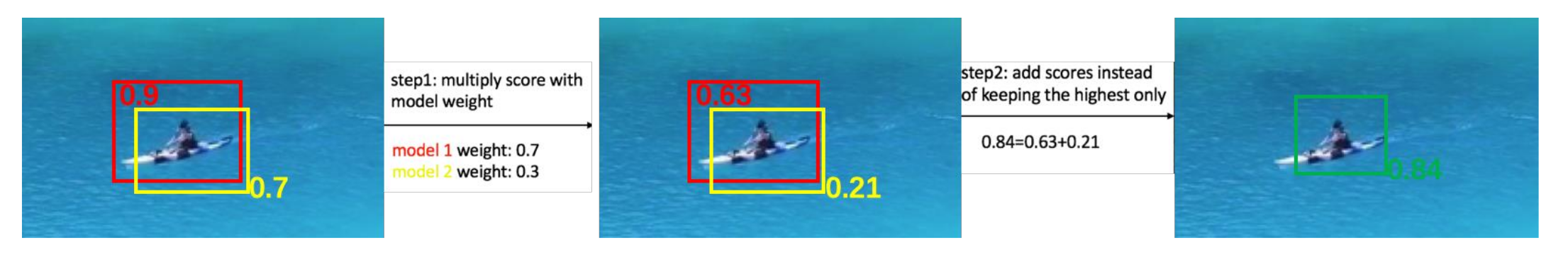}
\end{center}
   \caption{Illustration of hierarchical coarse-to-fine.}
\label{Fig:1 corseTofine}
\end{figure}
\noindent\textbf{Model Ensemble.} Authors further train networks with various backbones such as SENet-154\cite{hu2018squeeze},ResNet-200, CBNet\cite{liu2020cbnet} and Res2Net-200\cite{gao2019res2net} for combination. Existing ensemble methods can effectively fuse the networks with relatively close size and performance. However, the results get worse when it comes to models with very different size and performance, since the smaller models deteriorate the performance of the bigger ones. To handle this, authors propose a simple and effective ensemble method called hierarchical coarse-to-fine as shown in Fig 3.\newline
\subsection{Team STY-402}
\textbf{Cheng Gao, Wei Tang, Lizuo Jin}
(Southeast University)\newline
\begin{figure}[t]
\begin{center}
   \includegraphics[width=1.0\linewidth]{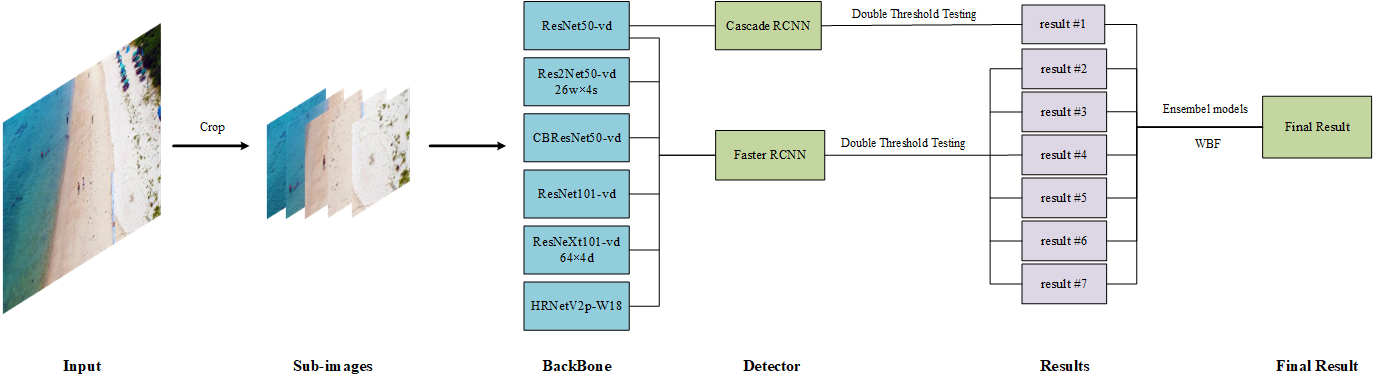}
\end{center}
   \caption{Illustration of pipline of Team STY-402.}
\label{Fig:7 backbone}
\end{figure}
\indent Authors adopt Faster R-CNN with ResNet-50, FPN, DCNv2 as the baseline model. All models are pre-trained on MS COCO. The baseline employs feature pyramid levels from P$_{2}$ to P$_{6}$ and the area of anchors are defined as (12$^{2}$,24$^{2}$,48$^{2}$,96$^{2}$,192$^{2}$) pixels. Deformable convolutions are applied in all convolution layers in stages 3-5 in ResNet-50.\newline
\textbf{Backbone.} In baseline, the weights of the first stage are frozen. Since the difference between MS COCO and TinyPerson is obvious, all the convolution layers in the backbone are unfreezed. In addition, Batch Normalization layer is added after each convolution layer of FPN. 
Authors replace ResNet with ResNet-vd. Moreover, Res2Net is a new multi-scale backbone architecture, which can further improve the performance of several representative computer vision tasks with no effort. 
Authors also train Res2Net-50 with 26w$\times$4s, and the performance improves by 3$\%$ compared to ResNet-50-vd.\newline
\textbf{Multi-scale Training.} The scale of short side is randomly sampled from {832, 896, 960, 1024, 1088, 1152, 1216, 1280, 1344, 1408, 1472, 1500} and the longer edge is fixed to 2000 in PaddleDetection. In particular, due to the limited GPU memory, the maximum value of the short side is set to 1280 when training ResNeXt101(64$\times$4d)\cite{xie2017aggregated}. In MMDetection, the scale of short side is randomly sampled from {480, 528, 576, 624, 672, 720, 768, 816, 912, 960} and the longer edge is fixed to 1280.\newline
\textbf{Training Tricks.} In the training stage, the number of proposals before NMS is changed from 2000 to 12000. And the data is changed to 6000 in testing stage.
\textbf{Data Augmentation.} Random horizontal flip, random cropping, random expanding and CutMix are adopt to augment training data. VisDrone is also used as additional data, which only use categories 0, 1, 2, and delete categories 3 - 11.\newline
\textbf{Large Scale Testing.} A large scale (1280 $\times$ 960) is adopted for testing. In order to get better performance, another large scale (1024 $\times$ 768) is also used for testing at the same time.\newline
\textbf{Double Threshold Testing.} If the highest confidence of a sub-image detection results is less than a certain threshold (0.5), the sub-image will be regarded as a pure background image and ignored.\newline
\noindent\textbf{Model Ensemble.} They train 7 models using different backbones as shown in Fig 4. Except HRNetV2p-W18 is trained on MMDetection, the rest of the models are trained on PaddleDetection. Finally, the final ensemble results is obtained by weighted boxes fusion (the IoU threshold is 0.6).

\subsection{Team BRiLliant}
\textbf{Mingbo Hong, Yuchao Yang, Huan Luo, Shuiwang Li, Qijun Zhao}\newline
(College of Computer Science, Sichuan University)\newline
\begin{figure}[t]
\begin{center}
   \includegraphics[width=0.7\linewidth]{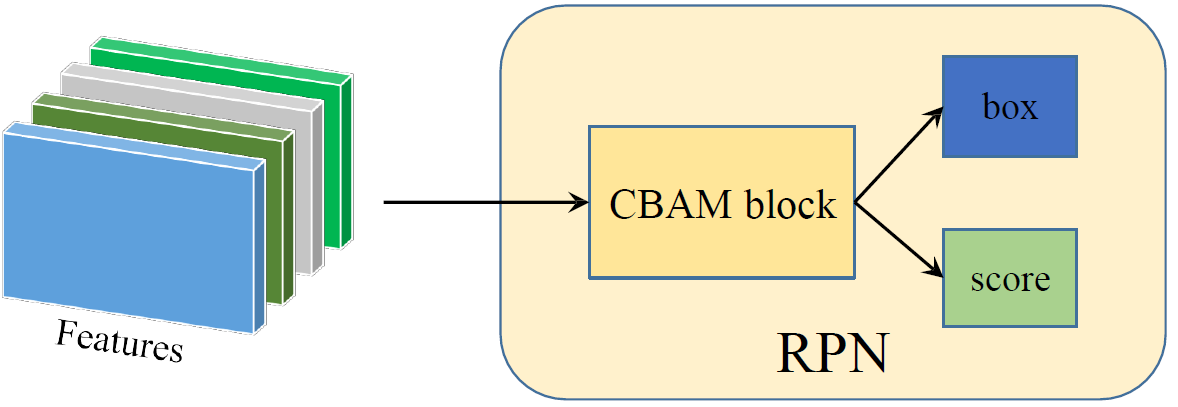}
\end{center}
   \caption{Illustration of improved CBAM in Region Proposal Network}
\label{Fig:CBAM}
\end{figure}
\indent To explore more detail features in tiny objects, authors utilize High Resolution Net (HRNet)\cite{sun2019deep} as backbone network, which allows the network to extract high-resolution representation. To simultaneously detect objects of varying scales, authors introduce an improved Convolutional Block Attention Module (CBAM)\cite{woo2018cbam} in Region Proposal Network to guide network to “Look Where” as shown in Fig 5 . Unlike the traditional CBAM, improved CBAM adds a suppression block to balance the attention value between objects of different scales. Furthermore, in order to raise different numbers of proposals for different scale objects, authors use “Top k sampler” instead of fixed threshold to select positive samples as shown in Fig 6, and the selection is based on the IOU metric rather than center distance that was utilized in ATSS\cite{zhang2020bridging}. The proposed sampler is adaptive to the scale of objects, which can be more accurate in detecting tiny objects, whereas ATSS may not generate any positive samples for tiny objects at all.

\begin{figure}[t]
\begin{center}
   \includegraphics[width=0.8\linewidth]{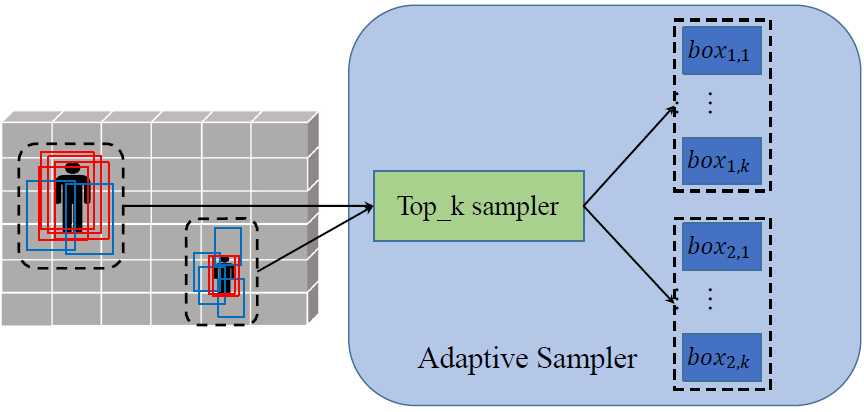}
\end{center}
   \caption{Illustration of Adaptive Sampler. The blue solid line indicates low-quality proposals, the red solid line indicates high-quality proposals, and Adaptive Sampler will adaptively select positive sample according to the quality of proposals.}
\label{Fig:adaptive sampler}
\end{figure}

\section{Conclusions}

We held the 1st TOD Challenge to encourage novel visual recognition research into tiny object detection. We used TinyPerson as the challenge dataset, and both AP and MR as evaluation metric.
Approximately 36 teams around the globe participated in this competition, in which top-3 leading teams, together with their methods, are briefly introduced in this paper. It is our vision that tiny object detection should extend far beyond person detection. Tiny object detection related tasks are important for many real-world computer vision applications, and the the advancement of addressing its technical challenges can help general object detection research as well. We hope our 1st TOD Challenge is a useful initial step in this promising research direction.



%
%
\bibliographystyle{splncs04}
\bibliography{egbib}
\end{document}